\documentclass[letterpaper]{article} 
\usepackage{aaai24}  
\usepackage{times}  
\usepackage{helvet}  
\usepackage{courier}  
\usepackage[hyphens]{url}  
\usepackage{graphicx} 
\usepackage{natbib}  
\usepackage{caption} 
\usepackage{algorithm}
\usepackage{algorithmic}

\usepackage{cuted}
\usepackage{amsfonts} 
\usepackage{amsmath}
\usepackage{array}
\usepackage{mathtools}
\usepackage{comment}
\usepackage{bm}
\usepackage{multicol}
\usepackage{multirow}
\usepackage{threeparttable}
\DeclareMathAlphabet{\mathsfit}{\encodingdefault}{\sfdefault}{m}{sl}
\SetMathAlphabet{\mathsfit}{bold}{\encodingdefault}{\sfdefault}{bx}{n}

\usepackage{color}
\usepackage{textcase}
\usepackage{bbding}
\usepackage{array}
\usepackage{booktabs}
\usepackage{enumitem}
\usepackage{arydshln}
\usepackage{makecell}
\usepackage{graphicx}
\usepackage{pifont}
\usepackage{soul}

\makeatletter
\renewcommand*\env@matrix[1][\arraystretch]{%
  \edef\arraystretch{#1}%
  \hskip -\arraycolsep
  \let\@ifnextchar\new@ifnextchar
  \array{*\c@MaxMatrixCols c}}
\makeatother

\usepackage{newfloat}
\usepackage{listings}
\DeclareCaptionStyle{ruled}{labelfont=normalfont,labelsep=colon,strut=off} 
\floatstyle{ruled}
\newfloat{listing}{tb}{lst}{}
\floatname{listing}{Listing}
\title{GCNext: Towards the Unity of Graph Convolutions for Human Motion Prediction}

\author {
    Xinshun Wang\textsuperscript{\rm 1,\rm 2},
    Qiongjie Cui\textsuperscript{\rm 3},
    Chen Chen\textsuperscript{\rm 4},
    Mengyuan Liu\textsuperscript{\rm 2}\thanks{Corresponding author: nkliuyifang@gmail.com}
}
\affiliations {
    \textsuperscript{\rm 1}School of Intelligent Systems Engineering, Sun Yat-sen University\\
    \textsuperscript{\rm 2}National Key Laboratory of General Artificial Intelligence, Peking University, Shenzhen Graduate School\\
    \textsuperscript{\rm 3}Xiaohongshu Inc.\\
    \textsuperscript{\rm 4}Center for Research in Computer Vision, University of Central Florida
}

\begin{document}

\maketitle

\begin{abstract}
    The past few years has witnessed the dominance of Graph Convolutional Networks (GCNs) over human motion prediction.
    Various styles of graph convolutions have been proposed, with each one meticulously designed and incorporated into a carefully-crafted network architecture.
    This paper breaks the limits of existing knowledge by proposing \textbf{Universal Graph Convolution} (\textbf{UniGC}), a novel graph convolution concept that re-conceptualizes different graph convolutions as its special cases.
    Leveraging UniGC on network-level, we propose \textbf{GCNext}, a novel GCN-building paradigm that dynamically determines the best-fitting graph convolutions both sample-wise and layer-wise. GCNext offers multiple use cases, including training a new GCN from scratch or refining a preexisting GCN.
    Experiments on Human3.6M, AMASS, and 3DPW datasets show that, by incorporating unique module-to-network designs, GCNext yields up to $9\times$ lower computational cost than existing GCN methods, on top of achieving state-of-the-art performance.
    Our code is available at https://github.com/BradleyWang0416/GCNext.
\end{abstract}

\section{Introduction}
\begin{figure*}[t]
      \centering
      \includegraphics[width=\textwidth]{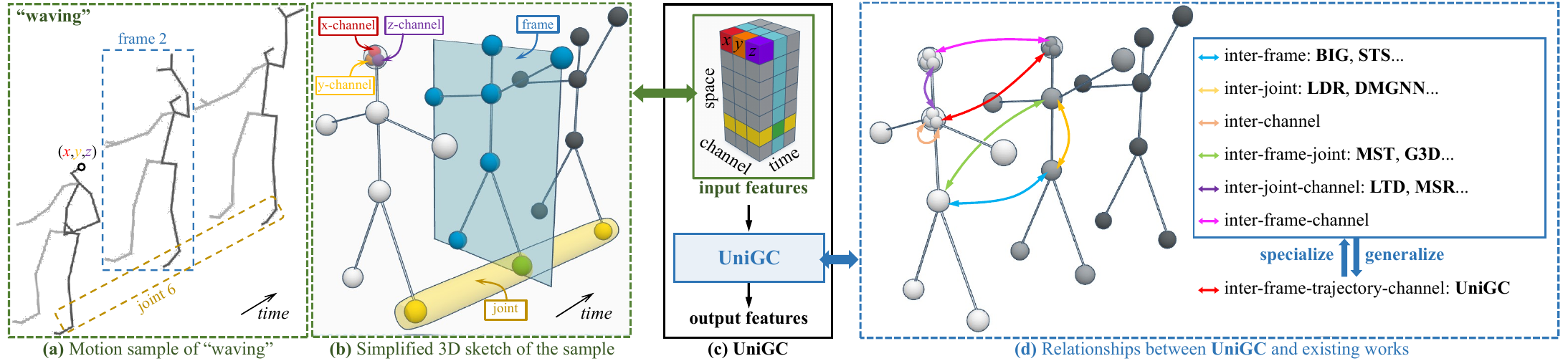}
      \caption{\small
      Our proposed Universal Graph Convolution (UniGC) extensively explores the inter-relationships within and across different aspects of human motion: space, time, and channel.
      (a) 3-frame snippet of a "waving" sample.
      (b) Simplified sketch where each frame of the sample contains 7 joints, and each joint contains 3 coordinates, namely, x-, y- and z-channel. (c) We propose UniGC to extract features from the given sample. (d) Graph convolutions for skeleton data in existing works such as BIG~\cite{ma2022progressively}, STS~\cite{sofianos2021space}, LDR~\cite{cui2020learning}, DMGNN~\cite{li2020dynamic}, MST~\cite{li2021multiscale}, G3D~\cite{liu2020disentangling}, LTD~\cite{mao2019learning}, MSR~\cite{dang2021msr}, DDGCN~\cite{wang2024dynamic} focus on only a subset of all human motion aspects, and can be seen as variants of UniGC's special cases.
      }
      \label{fig:teaser}
    \end{figure*}

The idea of predictable human motion has attracted much research interest over the years across a wide range of applications such as human-robot interaction and autonomous driving.
To generate plausible predictions, early successes were achieved with RNNs~\cite{fragkiadaki2015recurrent,ghosh2017learning} and CNNs~\cite{butepage2017deep,li2018convolutional}.
At present, human motion prediction is dominated almost exclusively by Graph Convolutional Networks (GCNs)~\cite{kipf2016semi}, due to the innate graph-like nature of human body joints and bones.
Human motion data are often in the form of 3D skeleton sequences, where each dimension corresponds to a different aspect---time, space, or channel, as shown in Fig. \ref{fig:teaser}.
To extract features, various graph convolutions are proposed. They either focus on aggregating information in a single dimension~\cite{li2020dynamic,cui2020learning}, or combine aggregations in two dimensions without considering the remaining one~\cite{wang2024dynamic,ma2022progressively,li2021multiscale,mao2019learning,sofianos2021space,liu2020disentangling}.
There are yet a number of other graph convolution types unexplored by existing works.
One is naturally motivated to ask: \textit{What makes an ideal type of graph convolution for human motion prediction}?

To answer this question, we propose \textbf{Universal Graph Convolution (UniGC)}, a novel graph convolution concept that pushes the boundaries of existing knowledge by re-conceptualizing different graph convolutions, both existing and unexplored, as its special cases.
In the most general case, UniGC uses 6D global adjacency to encode inter-relationships within and across space, time, and channels, which can be categorized into 7 types as shown in Fig. \ref{fig:teaser}.
The entry at position $\left[i,k,m,j,\ell,n\right]$ of the 6D adjacency represents the relationship between channel-$m$ of joint-$k$ at time-$i$ and channel-$n$ of joint-$\ell$ at time-$j$.
Using two techniques---adjacency masking and sub-adjacency tying, UniGC can be specialized into various different special cases, which focus on different subsets of all the 7 types of relationships. Existing graph convolutions are either the vanilla form or variations of these special cases, such as spatial~\cite{cui2020learning,li2020dynamic,ma2022progressively}, temporal~\cite{ma2022progressively}, spatial-temporal~\cite{li2021multiscale}, dense~\cite{ma2022progressively}, trajectory~\cite{mao2019learning,mao2020history,dang2021msr}, space-time-separable (STS)~\cite{sofianos2021space}, and unified (G3D)~\cite{liu2020disentangling} graph convolutions.
%
%
This gives rise to the second question: \textit{What is the best way for UniGC to be leveraged network-wise to better address the task}?

An intuitive answer based on common practices would be to stack multiple layers of UniGC. However, its 6D global adjacency would require too many parameters, making computation less efficient and optimization more challenging.
This paper gives a better answer by proposing \textbf{GCNext}, which exploits the versatility and adaptability of UniGC on network-level.
In contrast to existing approaches which typically involve manually choosing one from all possible types of graph convolutions, stacking it multiple layers, and evaluating the resulting network,
GCNext is structured to make the choice-making process learnable and sample-specific, where each layer dynamically chooses the optimal graph convolution type based on each different sample.
Specifically, each layer in GCNext contains several graph convolutional blocks of different types in parallel and a light-weight selector block tasked with choosing the best-fitting one for each sample.
Other network designs include a reinvention of update operations, and a replacement of batch normalization with layer normalization, both of which lead to performance improvement and efficiency boost.
Combining these designs, GCNext can yield up to $9\times$ reduction in computational cost compared to existing GCNs with comparable model sizes.
GCNext offers multiple use cases. First, it can be trained from scratch to obtain a GCN combining the strengths of different graph convolutions both layer-wise and sample-wise. Second, it can be used to refine a preexisting GCN with a predetermined type of graph convolution.

Our proposed UniGC and GCNext combined represent a module-to-network paradigm package, providing an effective approach that better addresses human motion prediction.
In summary, our contributions are three-fold:
\begingroup
\setlist[itemize,1]{leftmargin=0.37cm}
\begin{itemize}
    \item
    On concept-level, we propose UniGC, a novel graph convolution concept that re-conceptualizes different graph convolutions as its special cases. The specialization is achieved through two techniques: adjacency masking and sub-adjacency tying. UniGC represents not only a new concept of graph convolution, but also a new paradigm of designing graph convolutions.
    \item 
    On network-level, we propose GCNext, a novel GCN framework that dynamically determines the optimal GCN architecture both sample-wise and layer-wise. GCNext offers multiple use cases, including training a new GCN from scratch or refining a preexisting GCN. 
    \item
    We conduct extensive experiments on three benchmark datasets, Human3.6M, AMASS, and 3DPW, which show that GCNext can yield up to $9\times$ lower computational cost than existing GCNs with comparable model sizes, besides achieving state-of-the-art performance.
\end{itemize}
\endgroup

\section{Related Work}

\noindent\textbf{GCNs For Human Motion Data. }
Graph Convolutional Networks (GCNs)~\cite{kipf2016semi}, a popular type of graph neural network~\cite{scarselli2008graph,hamilton2017inductive,zaheer2017deep}, have wide-ranging applications in human motion prediction, action recognition~\cite{yan2018spatial,chen2021channel,shi2021adasgn,liu2023temporal,liu2022generalized} and so on.
Human motion data typically have three dimensions: space, time, and channel.
Spatial graph convolutions are employed to aggregate information in the spatial domain~\cite{cui2020learning,li2020dynamic}.
Temporal graph convolutions aggregate temporal information across different frames~\cite{ma2022progressively,li2021multiscale,sofianos2021space}.
Spatial-channel graph convolutions treat each trajectory as a node~\cite{mao2019learning}.
Spatial-temporal graph convolutions capture both spatial and temporal relationships~\cite{yan2018spatial,liu2020disentangling,zhong2022spatio,li2021multiscale}.
A related work to ours is G3D~\cite{liu2020disentangling}, which proposes a unified spatial-temporal graph convolution to aggregate cross-space-time features.
However, the inter-channel relationships, known to be important for achieving superior performance~\cite{mao2019learning,mao2020history,dang2021msr,guo2023back,mao2021multi}, is not addressed in G3D.
G3D can be seen as a variant of a special case in our UniGC.
Therefore, our UniGC is a more inclusive and unified concept, taking into account richer information with and across different spatial, temporal, and channel aspects of human motion.

\noindent\textbf{GCN-based Human Motion Prediction. }
LTD~\cite{mao2019learning} considers each joint coordinate as a node in the graph. Follow-up works incorporate techniques that better address the task, such as attention~\cite{mao2020history,mao2021multi}, and multi-scale graph~\cite{dang2021msr,li2020dynamic}.
LDR~\cite{cui2020learning} employ spatial graph convolutions to exploit skeleton structure, with temporal CNNs to learn temporal information.
Later works combine spatial and temporal graph convolutions to learn spatial-temporal dependencies~\cite{li2021multiscale,ma2022progressively,sofianos2021space}.
DDGCN~\cite{wang2024dynamic} proposes a dynamic dense graph convolution that unifies spatial and temporal domains and dynamically learns sample-specific dependencies.
In contrast to current approaches focusing on a subset of human motion aspects, our UniGC can provide both global modeling and more specific focus on local patterns among space, time, and channels, whose strengths can be combined by our GCNext.

    \renewcommand{\arraystretch}{1.2}
    \begin{table}[t]
    \centering
    
    \resizebox{\columnwidth}{!}{
    \begin{tabular}{m{2.9cm}|m{2.3cm}|m{3.1cm}} \hline
    \multicolumn{3}{c}{\textbf{Operations}} \\ \hline
    Notation & Variables & Description \\ \hline
    \multirow{2}{*}{$\mathbf{Y}=\mathcal{R}_{AB,CD}(\mathbf{X})$} & \multirow{2}{*}{\shortstack[l]{$\mathbf{X}:(A,B,C,D)$ \\ $\mathbf{Y}:(AB,CD)$}} & \multirow{2}{*}{array reshape} \\ 
    && \\ \hline
    \multirow{2}{*}{$\mathbf{Y}=\textit{G}(\mathbf{X};\mathbf{A})$} & \multirow{2}{*}{\shortstack[l]{$\mathbf{X}$, $\mathbf{Y}$: features \\ $\mathbf{A}$: adjacency}}  & \multirow{2}{*}{\shortstack[l]{graph convolution \\ on $\mathbf{X}$ depending on $\mathbf{A}$}} \\
    && \\ \hline
    \multirow{2}{*}{$\mathbf{X}\odot\mathbf{Y}$} & \multirow{2}{*}{\shortstack[l]{$\mathbf{X}$, $\mathbf{Y}$: arrays \\ of equal shape}} & \multirow{2}{*}{element-wise multiply} \\
    && \\ \hline
    \end{tabular}}
    
    \resizebox{\columnwidth}{!}{
    \begin{tabular}{m{2.9cm}|m{2.3cm}|m{3.1cm}}
    \multicolumn{3}{c}{\textbf{Symbols}} \\ \hline
    Notation & Shape & Description \\ \hline
    $\mathbb{X}$ &  $(T,J,C)$ & input sample \\ \hline
    $\mathbf{X}_t = \mathbb{X}\left[ t,:,: \right]$ & $(J,C)$ & pose features at time $t$\\ \hline
    $\mathbf{x}=\mathcal{R}_{TJC}(\mathbb{X})$ & $(TJC,1)$ & flattened $\mathbb{X}$ \\ \hline
    $\mathbf{x}_t=\mathcal{R}_{JC}(\mathbf{X}_t)$ & $(JC,1)$ & flattened $\mathbf{X}_t$ \\ \hline
    $\mathbb{A}$  & $(T,J,C,T,J,C)$ & global adjacency \\ \hline
    \multirow{2}{*}{$\mathbf{A}=\mathcal{R}_{TJC,TJC}(\mathbb{A})$} & \multirow{2}{*}{$(TJC,TJC)$} & \multirow{2}{*}{\shortstack[l]{global adjacency \\ (matrix)}} \\
    && \\ \hline
    $\mathbb{M}$ & $(T,J,C,T,J,C)$ & adjacency mask \\ \hline
    \end{tabular}}
    \vspace{-0.5em}
    \caption{\small Notations of operations and symbols.
    }
    \label{table: notation}
    \end{table}
    \renewcommand{\arraystretch}{1}

\section{Universal Graph Convolution (UniGC)}\label{section: universal graph convolution}

Our goal is to predict the future motion sequence given the history sequence.
Suppose that a motion sequence consists of $T$ frames, $J$ joints and $C$ coordinates (channels).
Each sample is represented by a 3D array $\mathbb{X}$ of shape $(T,J,C)$.
For better readability, we list the notations of operations and symbols used in the paper in Table \ref{table: notation}.

\paragraph{\textbf{The Most General Case.}}
We first present the most general case of UniGC.
Given a sample $\mathbb{X}$, we represent it as a global graph with $TJC$ nodes whose relationships are stored in global adjacency $\mathbb{A}$ of shape $(T,J,C,T,J,C)$.
The most general case of UniGC is defined as:
\begin{equation}\label{eq: global graph convolution}
\begin{split}
&\mathbb{Y} = \textit{G} (\mathbb{X};\mathbb{A})\\
&\begin{cases}
    \mathbf{x} = \mathcal{R}_{TJC}(\mathbb{X}) ;\\
    \mathbf{A} = \mathcal{R}_{TJC,TJC}(\mathbb{A}); \\
    \mathbf{y} \leftarrow \mathbf{A} \mathbf{x} ;\\
    \mathbb{Y} = \mathcal{R}_{T,J,C}(\mathbf{y}),
\end{cases}
\end{split}
\end{equation}
where the UniGC operation takes $\mathbb{X}$ as input and outputs $\mathbb{Y}$.
Without any specialization, global adjacency $\mathbb{A}$ is fully parameterized, defining a global aggregation of node features over a fully-connected graph where every node aggregates information from all the nodes including itself.

Consistently with the space, time, and channel dimensions and their combinations, UniGC can be specialized into $C_3^1+C_3^2=6$ graph convolution types: spatial-temporal ($\textit{G}^\text{st}$), spatial-channel ($\textit{G}^\text{sc}$), temporal-channel ($\textit{G}^\text{tc}$), spatial ($\textit{G}^\text{s}$), temporal ($\textit{G}^\text{t}$), and channel ($\textit{G}^\text{c}$) graph convolutions.

\paragraph{\textbf{Special Case 1: Spatial-Channel Graph Convolution.}}
Based on the most general case, an spatial-channel adjacency mask $\mathbb{M}^\text{sc}$ is introduced as an extra input to UniGC:
\begin{equation}\label{eq: spatial-channel adjacency mask}
\begin{split}
    \mathbb{M}^\text{sc} [t_1,:,:,t_2,:,:] = 
    \begin{cases}
        \mathbf{1} & \text{if } t_1 = t_2; \\
        \mathbf{0} & \text{if } t_1 \neq t_2.
    \end{cases}
\end{split}
\end{equation}
The mask $\mathbb{M}^\text{sc}$ is applied upon the global adjacency $\mathbb{A}$, giving the definition of \textit{spatial-channel graph convolution}:
\begin{equation}\label{eq: spatial-channel graph convolution}
\begin{split}
&\mathbb{Y} = \textit{G}^\text{sc} (\mathbb{X};\mathbb{A}, \mathbb{M}^\text{sc}) \\
&\begin{cases}
    \mathbf{x} = \mathcal{R}_{TJC}(\mathbb{X}) ;\\
    \mathbf{A} = \mathcal{R}_{TJC,TJC}(\mathbb{A}) ;\\
    \mathbf{M}^\text{sc} = \mathcal{R}_{TJC,TJC}(\mathbb{M}^\text{sc}) ;\\
    \mathbf{y} \leftarrow (\mathbf{A} \odot \mathbf{M}^\text{sc}) \mathbf{x} ;\\
    \mathbb{Y} = \mathcal{R}_{T,J,C}(\mathbf{y}).
\end{cases}
\end{split}
\end{equation}
Component-wise, the effect of $\mathbf{M}^\text{sc}$ is to keep only the main-diagonal blocks of shape $(JC,JC)$ in $\mathbf{A}$ such that:
\begin{equation}\label{eq: spatial-channel adjacency}
    \left[ \begin{array}{c} \mathbf{y}_1 \\ \hdashline[2pt/2pt] \mathbf{y}_2 \\ \hdashline[2pt/2pt]  \vdots \\ \hdashline[2pt/2pt] \mathbf{y}_T \end{array} \right]
    \leftarrow \left[ \begin{array}{c;{2pt/2pt}c;{2pt/2pt}c;{2pt/2pt}c}
    \mathbf{A}_{11} & \bm{0} & \cdots & \bm{0} \\\hdashline[2pt/2pt]
    \bm{0} & \mathbf{A}_{22} & \cdots & \bm{0} \\\hdashline[2pt/2pt]
    \vdots & \vdots & \ddots & \vdots \\\hdashline[2pt/2pt]
    \bm{0} & \bm{0} & \cdots & \mathbf{A}_{TT}
    \end{array} \right]
    \left[ \begin{array}{c} \mathbf{x}_1 \\ \hdashline[2pt/2pt] \mathbf{x}_2 \\ \hdashline[2pt/2pt]  \vdots \\ \hdashline[2pt/2pt] \mathbf{x}_T \end{array} \right],
\end{equation}
where $\mathbf{A}_{tt}=\mathcal{R}_{JC,JC} (\mathbb{A}\left[ t,:,:,t,:,: \right])$.
The mask $\mathbb{M}^\text{sc}$ nullifies inter-frame relationships, restricting the network to aggregating features only within each frame independently of other frames. For any time step $t$, we have:
\begin{equation}\label{eq: spatial-channel adjacency single frame}
    \mathbf{y}_t \leftarrow \mathbf{A}_{tt}\mathbf{x}_t.
\end{equation}
Such frame-wise independence allows one to apply graph convolution frame-by-frame, and then concatenate the outputs of each frame together for the final output.

The above graph convolution can be more constrained, by tying the sub-adjacencies of different frames:
\begin{equation}
    \mathbf{A}_\text{share} = \mathbf{A}_{11}=\mathbf{A}_{22}=\cdots=\mathbf{A}_{TT}.
\end{equation}
Using this inter-frame sub-adjacency tying, we can derive \textit{temporally-tied spatial-channel graph convolution}:
\begin{equation}\label{eq: temporally-tied spatial-channel graph convolution}
\begin{split}
&\mathbb{Y} = \textit{G}^{\text{sc}*} (\mathbb{X};\mathbb{A}) \\
&\begin{cases}
    \mathbf{X} = \mathcal{R}_{JC,T}(\mathbb{X}) ;\\
    \mathbf{A}_\text{share} = \mathcal{R}_{JC,JC}(\mathbb{A}\left[ 1,:,:,1,:,: \right]) ;\\
    \mathbf{Y} \leftarrow \mathbf{A}_\text{share} \mathbf{X} ;\\
    \mathbb{Y} = \mathcal{R}_{T,J,C}(\mathbf{Y}).
\end{cases}
\end{split}
\end{equation}
Similar to Eq. \ref{eq: spatial-channel adjacency} and Eq. \ref{eq: spatial-channel adjacency single frame}, the matrix multiplication step in Eq. \ref{eq: temporally-tied spatial-channel graph convolution} has component-wise representation as:
\begin{equation}\label{eq: temporally-tied spatial-channel adjacency}
\begin{split}
    \left[ \mathbf{y}_1 , \mathbf{y}_2 , \cdots , \mathbf{y}_T \right]
    &\leftarrow
    \mathbf{A}_\text{share}
    \left[  \mathbf{x}_1 , \mathbf{x}_2 , \cdots , \mathbf{x}_T \right] ; \\
    \mathbf{y}_t & \leftarrow \mathbf{A}_\text{share} \mathbf{x}_t.
\end{split}
\end{equation}

The \textit{spatial-channel adjacency masking} and the \textit{inter-frame sub-adjacency tying} techniques combined allow the module to capture frame-specific inter-spatial-channel information while benefiting from shared insights from similar patterns in different frames. Also, the size of adjacency parameters is reduced from $T\times J\times C \times T\times J\times C$ (as in $\mathbb{A}$) to a minimum of $JC \times JC$ (as in $\mathbf{A}_\text{share}$).

    \begin{figure*}[t]
        \centering
        \includegraphics[width=0.99\textwidth]{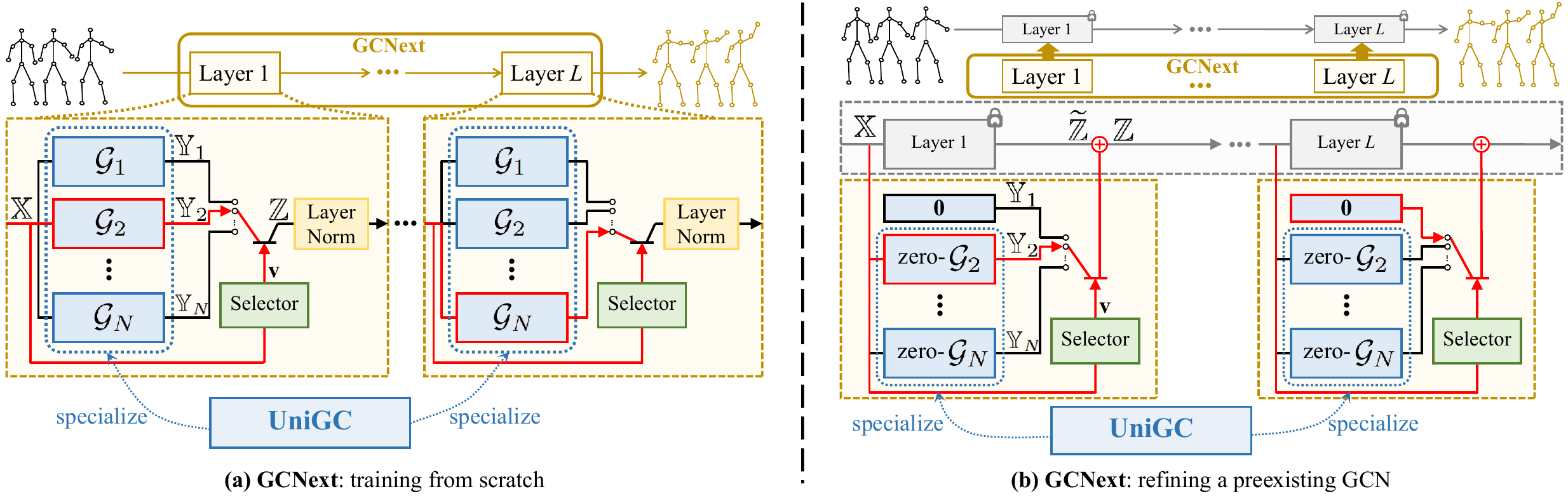}
        \caption{\small
        Two use cases of GCNext.
        (a) GCNext trained from scratch to dynamically determine the best-fitting GCN architecture at each layer for each sample.
        (b) GCNext as a plug-in to refine a preexisting GCN. By leveraging a novel ``zero-graph convolution'' scheme to initialize the adjacencies with zeros, GCNext preserves the qualities of the original GCN, and gradually refines it during training.
        }
        \label{fig: 2 network}
    \end{figure*}

\paragraph{\textbf{Special Case 2: Spatial-Temporal Graph Convolution.}}
Similar to special case 1, we introduce spatial-temporal adjacency mask $\mathbb{M}^\text{st}$, which is defined as:
\begin{equation}\label{eq: spatial-temporal adjacency mask}
\begin{split}
    \mathbb{M}^\text{st} [:,:,c_1,:,:,c_2] = 
    \begin{cases}
        \mathbf{1} & \text{if } c_1 = c_2 ;\\
        \mathbf{0} & \text{if } c_1 \neq c_2.
    \end{cases}
\end{split}
\end{equation}
The \textit{spatial-temporal graph convolution} $\textit{G}^{\text{st}}$ can be obtained by simply replacing the mask term $\mathbb{M}^\text{sc}$ in Eq. \ref{eq: spatial-channel graph convolution} with $\mathbb{M}^\text{st}$.
Also similar to special case 1, it can be further constrained by inter-channel sub-adjacency tying, becoming \textit{channel-tied spatial-temporal graph convolution}:
\begin{equation}\label{eq: channel-tied spatial-temporal graph convolution}
\begin{split}
&\mathbb{Y} = \textit{G}^{\text{st}*} (\mathbb{X};\mathbb{A}) \\
&\begin{cases}
    \mathbf{X} = \mathcal{R}_{TJ,C}(\mathbb{X}) ;\\
    \mathbf{A}_\text{share} = \mathcal{R}_{TJ,TJ}(\mathbb{A}\left[ :,:,1,:,:,1 \right]) ;\\
    \mathbf{Y} \leftarrow \mathbf{A}_\text{share} \mathbf{X} ;\\
    \mathbb{Y} = \mathcal{R}_{T,J,C}(\mathbf{Y}).
\end{cases}
\end{split}
\end{equation}

\paragraph{\textbf{Special Case 3: Spatial Graph Convolution.}}
Spatial graph convolution $\textit{G}^{\text{s}}(\mathbb{X};\mathbb{A})$ uses the spatial adjacency mask $\mathbb{M}^\text{s}=\mathbb{M}^\text{sc}\odot\mathbb{M}^\text{st}$, where only the spatial relationships within the same frame and the same channel are aggregated.

The rest of the special cases, including temporal-channel ($\textit{G}^{\text{tc}}$), temporal ($\textit{G}^{\text{t}}$), and channel ($\textit{G}^{\text{c}}$) graph convolutions, can all be derived in ways similar to special case 1--3.

\paragraph{\textbf{Connections with Existing Graph Convolutions.}}
Graph convolutions proposed by existing works largely are the vanilla form or are variants of UniGC's special cases.
For example, the \textit{temporally-tied spatial-channel graph convolution} (Eq. \ref{eq: temporally-tied spatial-channel graph convolution}) is used in LTD~\cite{mao2019learning}, while its variants are used in MSR-GCN~\cite{dang2021msr} and HisRep~\cite{mao2020history}.
The well-accepted concept of spatial-temporal graph convolutions in the community~\cite{ma2022progressively,li2021multiscale,sofianos2021space,liu2020disentangling} is actually the variation of the \textit{channel-tied spatial-temporal graph convolution} (Eq. \ref{eq: channel-tied spatial-temporal graph convolution}), while the channel-untied case is unexplored by existing work.

\paragraph{\textbf{Further Discussion.}}
It is worth noting that, while the formulas for UniGC appear complex in matrix form, they can be easily implemented using tensor operations, with the help of Einstein notation.

\section{GCNext Framework} \label{section: gchmp}
The most general case of UniGC requires a large size of parameters, making computation less efficient and optimization more challenging. To balance effectiveness and efficiency, we propose GCNext, which leverages the adaptability of UniGC, and offers multiple use cases.

    \begin{table*}[t]
    \centering
    \resizebox{\textwidth}{!}{
    \begin{tabular}{p{5.8cm}|p{1.4cm}|p{1cm}p{1cm}p{1cm}p{1cm}p{1cm}p{1cm}p{1cm}p{1cm}} \Xhline{0.3ex}
    \multirow{2}{*}{Model} & \multirow{2}{*}{Venue}  & \multicolumn{8}{c}{Mean Per Joint Position Error (mm)} \\
    &  & 80ms & 160ms & 320ms & 400ms & 560ms & 720ms & 880ms & 1000ms  \\ \Xhline{0.3ex}
    LTD~\cite{mao2019learning}& ICCV'19   	&11.2 	&23.4 	&47.9 	&58.9 	&78.3 	&93.3 	&106.0 	&114.0 	\\
    HisRep~\cite{mao2020history}&ECCV'20	&10.4 	&22.6 	&47.1 	&58.3 	&77.3 	&91.8 	&104.1 	&112.1  	 \\
    MSR-GCN~\cite{dang2021msr}&ICCV'21	 &11.3 	&24.3 	&50.8 	&61.9 	&80.0 	&-	&-	&112.9 	 \\
    PGBIG~\cite{ma2022progressively}&CVPR'22		 &10.6 	&23.1 	&47.1 	&57.9 	&76.3 	&90.7 	&102.4 	&109.7 \\
    siMLPe~\cite{guo2023back}&WACV'23 	&9.6 	&21.7 	&46.3 	&57.3 	&75.7 	&90.1 	&101.8 	&109.4  	\\
    \textbf{Ours}&    &\textbf{\textbf{9.3}}	&\textbf{\textbf{21.5}}	&\textbf{\textbf{45.5}}	&\textbf{\textbf{56.4}}	&\textbf{\textbf{74.7}}	&\textbf{\textbf{88.9}}	&\textbf{\textbf{100.8}}	&\textbf{\textbf{108.7}}  
    \\ \hline
    STSGCN~\cite{sofianos2021space}$^\dagger$ &ICCV'21 	&10.1&17.1&33.1&38.3&50.8&60.1&68.9&75.6    \\
    GAGCN~\cite{zhong2022spatio}$^\dagger$ &CVPR'22  &10.1&16.9&32.5&38.5&50.0&-&-&72.9 	  \\
    MotionMixer~\cite{bouazizi2022motionmixer}$^\dagger$ &IJCAI'22  &9.0&13.2&26.9&33.6&46.1&56.5&65.7&71.6   \\
    \textbf{Ours}$^\dagger$ &     &\textbf{\textbf{6.9}}	&\textbf{\textbf{12.7}}	&\textbf{\textbf{24.7}}	&\textbf{\textbf{30.5}}	&\textbf{\textbf{41.3}}	&\textbf{\textbf{50.7}}	&\textbf{\textbf{59.0}}	&\textbf{\textbf{64.7}}  	
    \\ \Xhline{0.3ex}
    \end{tabular}}
    \vspace{-0.5em}
    \caption{\small \textbf{Results on Human3.6M} at different prediction time steps in terms of mean per joint position error (MPJPE) in millimeter. 256 samples are tested for each action. $\dagger$ indicates methods that compute the average error over all frames, whose results are taken from the paper by \citet{bouazizi2022motionmixer}. Otherwise, the methods are evaluated at each frame, whose results are taken from the paper by \citet{guo2023back}. Our approach achieves better performance than others under both evaluation protocols.
    }
    \vspace{-0.5em}
    \label{h36m}
    \end{table*}

    \begin{table*}[t]
    \centering
    \resizebox{\textwidth}{!}{
    \begin{tabular}{p{2.3cm}|p{0.7cm}p{0.7cm}p{0.7cm}p{0.7cm}p{0.7cm}p{0.7cm}p{0.7cm}p{0.7cm}|p{0.7cm}p{0.7cm}p{0.7cm}p{0.7cm}p{0.7cm}p{0.7cm}p{0.7cm}p{0.7cm}} \Xhline{0.3ex}
    Dataset & \multicolumn{8}{c|}{AMASS-BMLrub} & \multicolumn{8}{c}{3DPW} \\ \hline
    milliseconds & 80 & 160 & 320 & 400 & 560 & 720 & 880 & 1000 & 80 & 160 & 320 & 400 & 560 & 720 & 880 & 1000 \\ \Xhline{0.3ex}
    LTD 	&11.0 	&20.7 	&37.8 	&45.3 	&57.2 	&65.7 	&71.3 	&75.2 	&12.6 	&23.2 	&39.7 	&46.6 	&57.9 	&65.8 	&71.5 	&75.5 	\\
    HisRep	&11.3 	&20.7 	&35.7 	&42.0 	&51.7 	&58.6 	&63.4 	&67.2 	&12.6 	&23.1 	&39.0 	&45.4 	&56.0 	&63.6 	&69.7 	&73.7 	\\
    siMLPe  &10.8 	&19.6 	&34.3 	&40.5 	&50.5 	&57.3 	&62.4 	&65.7 	&12.1 	&22.1 	&38.1 	&44.5 	&54.9 	&62.4 	&68.2 	&72.2 	\\ \hline
    Ours
    &\textbf{\textbf{10.2 }}	&\textbf{\textbf{19.3 }}	&\textbf{\textbf{34.1}} 	&\textbf{\textbf{40.3 }}	&50.6	&\textbf{\textbf{57.3 }}	&\textbf{\textbf{62.0 }}	&\textbf{\textbf{65.3 }}	&\textbf{\textbf{11.8 }}	&\textbf{\textbf{22.0} }	&\textbf{\textbf{37.9 }}	&\textbf{\textbf{44.2 }}	&55.1	&\textbf{\textbf{62.1 }}	&\textbf{\textbf{67.8 }} & \textbf{\textbf{72.0 }}
    \\ \Xhline{0.3ex}
    \end{tabular}}
    \vspace{-0.5em}
    \caption{\small  \textbf{Results on AMASS and 3DPW}. The results of other methods are taken from the paper by \citet{guo2023back}. 
    }
    \vspace{-0.5em}
    \label{amass 3dpw}
    \end{table*}

\paragraph{\textbf{Use Case 1: Training from Scratch.}}
As shown in Fig.~\ref{fig: 2 network}~(a), GCNext contains $L$ dynamic layers, where each layer has $N$ candidate graph convolution operations, a selector, and a layer normalization.
The selector is tasked with determining the optimal graph convolution operation for this layer based on the layer input $\mathbb{X}$.
Each of the $N$ candidates perform a different type of graph convolution:
\begin{equation}\label{eq: candidate graph convolution block}
\begin{split}
    \mathbb{Y}_i = \textit{G}_i (\mathbb{X} ; \mathbb{A}, \mathbb{M}_i, \theta_i), \forall i\in\left[1,N\right],
\end{split}
\end{equation}
where $\textit{G}_i \in \{ \textit{G}^\text{st},\textit{G}^\text{sc},\textit{G}^\text{tc},\textit{G}^\text{s},\textit{G}^\text{t},\textit{G}^\text{c} \}$, $\theta_i$ is the parameters that parameterize $\mathbb{A}$, and $\mathbb{M}_i$ is the adjacency mask required for specializing to $\textit{G}_i$.
The layer input $\mathbb{X}$ is also passed through the selector, which relies on a learnable process $S$ followed by Gumbel Softmax~\cite{jang2016categorical}:
\begin{equation}
\begin{split}
    \mathbf{v} = \textit{GumbelSoftmax} \big(\mathcal{S}(\mathbb{X};\theta_\text{s}) \big),
\end{split}
\end{equation}
where $\mathbf{v} = \left[ v_1,v_2,\cdots,v_N \right] \in \{0,1\}^N$ is a one-hot vector that indicates the optimal block, and $\mathcal{S}:\mathbb{R}^{T\times J\times C}\to\mathbb{R}^{N}$ is implemented with an average pooling followed by an MLP, which learns to assign higher values to the best-matched candidate.
The graph convolution output $\mathbb{Z}$ is obtained by $\mathbb{Z} = \sum_{i=1}^N v_i \mathbb{Y}_i$, followed by layer normalization to produce the final output of this layer.
During inference, the network directly passes the input through the optimal graph convolution block based on the index of the one-hot vector, thereby reducing computational cost.

    \begin{table*}[t]
    \centering
    \resizebox{\textwidth}{!}{
    \begin{tabular}{p{2.3cm}|p{0.7cm}p{0.7cm}p{0.7cm}p{0.7cm}|p{0.7cm}p{0.7cm}p{0.7cm}p{0.7cm}|p{0.7cm}p{0.7cm}p{0.7cm}p{0.7cm}|p{0.7cm}p{0.7cm}p{0.7cm}p{0.7cm}} \Xhline{0.3ex}
    \multirow{2}{*}{Model}    & \multicolumn{4}{c|}{Walking} & \multicolumn{4}{c|}{Eating} &  \multicolumn{4}{c|}{Smoking} & \multicolumn{4}{c}{Discussion} \\
        & 80 & 400 & 560 & 1000 & 80 & 400 & 560 & 1000 & 80 & 400 & 560 & 1000 & 80 & 400 & 560 & 1000 \\ \Xhline{0.3ex}
    LTD	&11.1 	&42.9 	&53.1 	&70.7 	&7.0 	&37.3 	&51.1 	&78.6 	&7.5 	&37.5 	&49.4 	&71.8 	&10.8 	&65.8 	&88.1 	&121.6 	\\
    HisRep	&10.0 	&39.8 	&47.4 	&58.1 	&6.4 	&36.2 	&50.0 	&75.7 	&7.0 	&36.4 	&47.6 	&69.5 	&10.2 	&65.4 	&86.6 	&119.8 	\\
    MSR-GCN	&10.8 	&42.4 	&53.3 	&63.7 	&6.9 	&36.0 	&50.8 	&75.4 	&7.5 	&37.5 	&50.5 	&72.1 	&10.4 	&65.0 	&87.0 	&116.8 	\\
    PGBIG 	&11.2 	&42.8 	&49.6 	&58.9 	&6.5 	&36.8 	&50.0 	&74.9 	&7.3 	&37.5 	&48.8 	&69.9 	&10.2 	&64.4 	&86.1 	&116.9 	\\
    siMLPe 	&9.9 	&39.6 	&46.8 	&55.7 	&5.9 	&36.1 	&49.6 	&74.5 	&6.5 	&36.3 	&47.2 	&69.3 	&9.4 	&64.3 	&85.7 	&116.3 	\\  
    \textbf{Ours} &\textbf{\textbf{8.8}}	&\textbf{\textbf{38.9}}	&\textbf{\textbf{46.4}}	&\textbf{\textbf{55.0}}	&\textbf{\textbf{5.9}}	&\textbf{\textbf{35.0}}	&\textbf{\textbf{48.2}}	&\textbf{\textbf{73.9 }}	&\textbf{\textbf{5.6}}	&\textbf{\textbf{36.1}}	&\textbf{\textbf{46.9}}	&\textbf{\textbf{68.6}}	&\textbf{\textbf{8.8}}	&\textbf{\textbf{63.1}}	&\textbf{\textbf{84.0}}	&\textbf{\textbf{114.8}}
    \\ \hdashline
    STSGCN$^\dagger$	&10.7 	&38.2 	&40.6	&51.8 	&6.7 	&31.6 	&33.9	&52.5 	&7.1 	&30.6 	&33.6	&50.1 	&9.7 	&45.0 	&53.4	&78.8 	\\
    GAGCN$^\dagger$	&10.3 	&32.4 	&39.9	&51.1 	&6.4 	&25.2 	&31.8	&51.4 	&7.1 	&24.3 	&31.1	&48.7 	&9.7 	&38.9 	&53.1	&76.9 	\\
    MotionMixer$^\dagger$	&7.3 	&28.6 	&-	&49.2 	&4.3 	&20.9 	&-	&47.4 	&4.7 	&21.4 	&-	&45.4 	&6.4 	&35.5 	&-	&78.0 	\\ 
    \textbf{Ours}$^\dagger$ &\textbf{\textbf{6.8}}	&\textbf{\textbf{24.3}}	&\textbf{\textbf{30.3}}	&\textbf{\textbf{40.3}}	&\textbf{\textbf{4.3}}	&\textbf{\textbf{18.9}}	&\textbf{\textbf{25.9}}	&\textbf{\textbf{42.6}}	&\textbf{\textbf{4.5}}	&\textbf{\textbf{19.8}}	&\textbf{\textbf{26.5}}	&\textbf{\textbf{41.2}}	&\textbf{\textbf{6.2}}	&\textbf{\textbf{32.9}}	&\textbf{\textbf{45.4}}	&\textbf{\textbf{70.8}}
    \\
    \end{tabular}
    }
    \resizebox{\textwidth}{!}{
    \begin{tabular}{p{2.3cm}|p{0.7cm}p{0.7cm}p{0.7cm}p{0.7cm}|p{0.7cm}p{0.7cm}p{0.7cm}p{0.7cm}|p{0.7cm}p{0.7cm}p{0.7cm}p{0.7cm}|p{0.7cm}p{0.7cm}p{0.7cm}p{0.7cm}} \Xhline{0.3ex}
    \multirow{2}{*}{Model}    & \multicolumn{4}{c|}{Waiting} & \multicolumn{4}{c|}{Walking Dog} &  \multicolumn{4}{c|}{Walking Together} & \multicolumn{4}{c}{Average} \\
     & 80 & 400 & 560 & 1000 & 80 & 400 & 560 & 1000 & 80 & 400 & 560 & 1000 & 80 & 400 & 560 & 1000 \\ \Xhline{0.3ex}
    LTD	&9.2 	&54.4 	&73.4 	&107.5 	&20.9 	&86.6 	&109.7 	&150.1 	&9.6 	&44.0 	&55.7 	&69.8 	&11.2 	&58.9 	&78.3 	&114.0 	\\
    HisRep	&8.7 	&54.9 	&74.5 	&108.2 	&20.1 	&86.3 	&108.2 	&146.9 	&8.9 	&41.9 	&52.7 	&64.9 	&10.4 	&58.3 	&77.3 	&112.1 	\\
    MSR-GCN	&10.4 	&62.4 	&74.8 	&105.5 	&24.9 	&112.9 	&107.7 	&145.7 	&9.2 	&43.2 	&56.2 	&69.5 	&11.3 	&61.9 	&80.0 	&112.9 	\\
    PGBIG 	&8.7 	&53.6 	&71.6 	&103.7 	&20.4 	&84.6 	&105.7 	&145.9 	&8.9 	&43.8 	&54.4 	&64.6 	&10.6 	&57.9 	&76.3 	&109.7 	\\
    siMLPe &7.8 	&53.2 	&71.6 	&104.6 	&18.2 	&83.6 	&105.6 	&141.2 	&8.4 	&41.2 	&50.8 	&61.5 	&9.6 	&57.3 	&75.7 	&109.4 	\\
    \textbf{Ours} &\textbf{\textbf{7.7}}	&\textbf{\textbf{52.6}}	&\textbf{\textbf{71.5}}	&\textbf{\textbf{104.0}}	&18.4	&\textbf{\textbf{82.8}}	&\textbf{\textbf{104.6}}	&142.8	&\textbf{\textbf{8.4}}	&\textbf{\textbf{40.5}}	&\textbf{\textbf{50.2}}	&\textbf{\textbf{60.8}}	&\textbf{\textbf{9.3}}	&\textbf{\textbf{56.4}}	&\textbf{\textbf{74.7}}	&\textbf{\textbf{108.7}}
    \\ \hdashline
    STSGCN$^\dagger$	&8.6 	&40.7 	&47.3	&72.0 	&17.6 	&66.4 	&74.7	&102.6 	&8.6 	&35.1 	&38.9	&51.1 	&10.1 	&38.3 	&51.7	&75.6 	\\
    GAGCN$^\dagger$	&8.5 	&33.8 	&45.9	&69.3 	&17.0 	&59.4 	&70.1	&91.3 	&-	&-	&-	&-	&10.1 	&38.5 	&50.0	&72.9 	\\
    MotionMixer$^\dagger$	&5.4 	&30.0 	&-	&68.2 	&13.4 	&54.1 	&-	&99.6 	&5.9 	&27.4 	&-	&50.4 	&9.0 	&33.6 	&-	&71.6 	\\ 
    \textbf{Ours}$^\dagger$	&\textbf{\textbf{5.4}}	&\textbf{\textbf{27.4}}	&\textbf{\textbf{38.0}}	&\textbf{\textbf{61.3}}	&\textbf{\textbf{13.3}}	&\textbf{\textbf{49.0}}	&\textbf{\textbf{62.6}}	&\textbf{\textbf{89.8}}	&6.1	&\textbf{\textbf{23.4}}	&\textbf{\textbf{30.1}}	&\textbf{\textbf{41.6}}	&\textbf{\textbf{6.9}}	&\textbf{\textbf{30.5}}	&\textbf{\textbf{41.3}}	&\textbf{\textbf{64.7}}
    \\ \Xhline{0.3ex}
    \end{tabular}
    }
    \caption{\small \textbf{Action-wise results on Human3.6M}.
    }
    \label{h36m per action}
    \end{table*}

    \begin{table}[t]
    \centering
    \resizebox{0.99\columnwidth}{!}{
    \begin{tabular}{m{2cm}|l|l|l} \Xhline{0.3ex}
    Model refined  & 80 &560 & 1000  \\   \Xhline{0.3ex}
    LTD  &11.2&78.3&114.0   \\
    Ours+LTD          &\textbf{9.9} ($\downarrow 1.3$)
    &\textbf{77.6} ($\downarrow0.7$)
    &\textbf{110.5} ($\downarrow3.5$)\\ \hdashline
    MSR-GCN  &11.3&80.0&112.9\\
    Ours+MSR      & \textbf{10.5} ($\downarrow 0.8$)
    &\textbf{78.1} ($\downarrow1.9$)
    &\textbf{110.8} ($\downarrow2.1$)\\ \hdashline
    PGBIG &10.6&76.3&109.7\\
    Ours+PGBIG        &\textbf{9.7} ($\downarrow 0.9$)
    &\textbf{75.7} ($\downarrow0.6$)
    &\textbf{108.8} ($\downarrow0.9$)  \\
    \Xhline{0.3ex}   
    \end{tabular}}
    \caption{\small \textbf{Results on Use Case 2}: Refining preexisting GCNs.
    }
    \label{table refine}
    \end{table}

\paragraph{\textbf{Use Case 2: Refining Preexisting GCNs.}}

Alternatively, GCNext can be used to refine a preexisting GCN.
Suppose that the preexisting GCN is built on graph convolution $\textit{G}_\tau$. Fig. \ref{fig: 2 network} (b) shows the case of $\tau=1$. At each layer, the network decides whether or not it needs to be refined, and if so, which graph convolution operation to use to refine it.
Let $\widetilde{\mathbb{Z}}=\textit{G}_\tau (\mathbb{X} ; \mathbb{A}, \mathbb{M}_\tau)$ denote the locked original output of this layer in the preexisting GCN. We re-define Eq. \ref{eq: candidate graph convolution block} as:
\begin{equation}\label{eq: refine candidate graph convolution block}
\begin{split}
    \mathbb{Y}_i = \begin{cases}
        \bm{0}, &\text{if } i=\tau;  \\
        \textit{G}_i (\mathbb{X} ; \mathbb{A}, \mathbb{M}_i, \theta_i), &\text{if } i\neq\tau.
    \end{cases}
    \quad \forall i\in\left[1,N\right]
\end{split}
\end{equation}
The graph convolution output $\mathbb{Z}$ is obtained as:
\begin{equation}
    \mathbb{Z} = \widetilde{\mathbb{Z}} + \sum_{i=1}^N v_i \mathbb{Y}_i.
\end{equation}
During training, the adjacencies $\mathbb{A}$ are initialized at zeros. Therefore, in the first train step, we have:
\begin{equation}
    \begin{cases}
        \theta_i=\bm{0}, \forall i\in\left[1,N\right], i\neq\tau ;\\
        \textit{G}_i (\mathbb{X} ; \mathbb{A}, \mathbb{M}_i, \theta_i)=\bm{0}, \forall i\in\left[1,N\right], i\neq\tau  ;\\
        \mathbb{Z} = \widetilde{\mathbb{Z}}.
    \end{cases}
\end{equation}
It indicates that, before optimization, the network is the same as the preexisting GCN as if GCNext did not exist.
The qualities of that GCN are perfectly preserved. Therefore, further optimization takes effect in a fine tuning manner.

    \begin{table}[t]
    \centering
    \resizebox{0.99\columnwidth}{!}{
    \begin{tabular}
    {m{1.8cm}|l|l|l|l} \Xhline{0.3ex}
    \multirow{2}{*}{Model}     & \multicolumn{2}{c|}{FLOPs (M)}    & \multirow{2}{*}{\shortstack[l]{Inference\\time (ms)}} & \multirow{2}{*}{Memory (GB)} \\   \cline{2-3}
    &train&inference&& \\ \Xhline{0.3ex}
    MSR-GCN   & 1449.6 & 1449.6  & 3.9   & 5.1 \\
    PGBIG     & 224.8 & 224.8   & 2.1   & 2.5 \\ 
    HisRep    & 148.7 & 148.7   & 1.6   & 2.3 \\
    LTD       & 133.3 & 133.3   & 1.2   & 2.2  \\ \hline
    Ours   & 29.1 & 14.5   &  0.8 & 3.1\\  \Xhline{0.3ex}
    \end{tabular}}
    \caption{\small \textbf{Comparison of efficiency} between existing GCNs and our \textbf{GCNext} in terms of train and inference FLOPs, inference time, and memory. Our GCNext is generally more efficient.
    }
    \label{table efficiency}
    \end{table}

\paragraph{\textbf{Other Network Designs.}}
We adopt a few other network designs to balance effectiveness and efficiency.

1) We reinvent the \textit{update} operations in traditional GCNs. The update operation usually uses a weight matrix to transform input features into high-dimensional hidden features, which is the main source of computation cost in GCNs. At each layer of GCNext, we employ one \textit{dimension-preserving update} operation following the selection step, which saves a lot of computation cost, compared to updating high-dimensional features in every graph convolution.

2) We replace \textit{batch normalization}, commonly applied in traditional GCNs, with \textit{layer normalization}, which requires fewer parameters and computes more efficiently.

\section{Experiments}
We report the Mean Per Joint Position Error (MPJPE), which is the preferred metric in human motion prediction.
The lower the MPJPE, the better the performance.

\subsection{Implementation Details}
The model was trained on RTX 3080 Ti GPU, and the training consumed about 3GB memory and took about 3 hours with batch size of 256. Testing took about 13 seconds. For H3.6M dataset, the model is trained for 85k iterations. The learning rate starts with 0.0006, and drops to 0.000005 after 75k iterations. For AMASS dataset, the model is trained for 115k iterations. The learning rate starts with 0.0003, and drops to 0.000001 after 100k iterations.

\subsection{Datasets}
\noindent\textbf{Human3.6M (H3.6M)}~\cite{ionescu2013human3} comprises 15 types of actions by 7 actors. 
Following siMLPe~\cite{guo2023back}, the dataset is preprocessed and converted into 3D coordinates, and each pose contains 22 joints. For the validation and test sets, we respectively use subject 11 and subject 5. The remaining 5 subjects for training.

\noindent\textbf{AMASS}~\cite{mahmood2019amass} combines multiple Mocap datasets unified by SMPL parameterization. Following siMLPe~\cite{guo2023back}, we use AMASS-BMLrub as the test set and split the rest into training and validation sets.

\noindent\textbf{3D Pose in the Wild (3DPW)}~\cite{von2018recovering} includes activities captured from indoor and outdoor scenes. We evaluate 18 joints using the model trained on AMASS.

\subsection{Comparison with State-of-the-Art Methods}

\noindent\textbf{Comparison on Human3.6M. }
Considering that the baseline methods follow two separate testing protocols: error computation at each time step~\cite{mao2020history} and error averaging~\cite{sofianos2021space}, we test our method on both protocols. Table \ref{h36m} shows the results on H3.6M using average Mean Per Joint Position Errors (MPJPEs) at different time steps for all actions. Our method surpasses all others under both testing protocols.
Table \ref{h36m per action} shows action-wise results for 7 exemplary actions in H3.6M.
Table \ref{table refine} reports the results of existing GCNs refined with GCNext, showing that refining with GCNext brings performance boost.
Table \ref{table efficiency} compares the efficiency of different GCNs, showing that GCNext is generally more efficient.

\noindent\textbf{Comparison on AMASS and 3DPW. }
In Table \ref{amass 3dpw}, we present the results obtained from AMASS and 3DPW datasets. As there are two distinct evaluation protocols~\cite{mao2020history,sofianos2021space}, we evaluate our method under both. The models are trained on AMASS and tested on AMASS-BMLrub and 3DPW.

\subsection{Ablation Studies}
\begin{figure}[t]
    \centering
    \includegraphics[width=0.9\columnwidth]{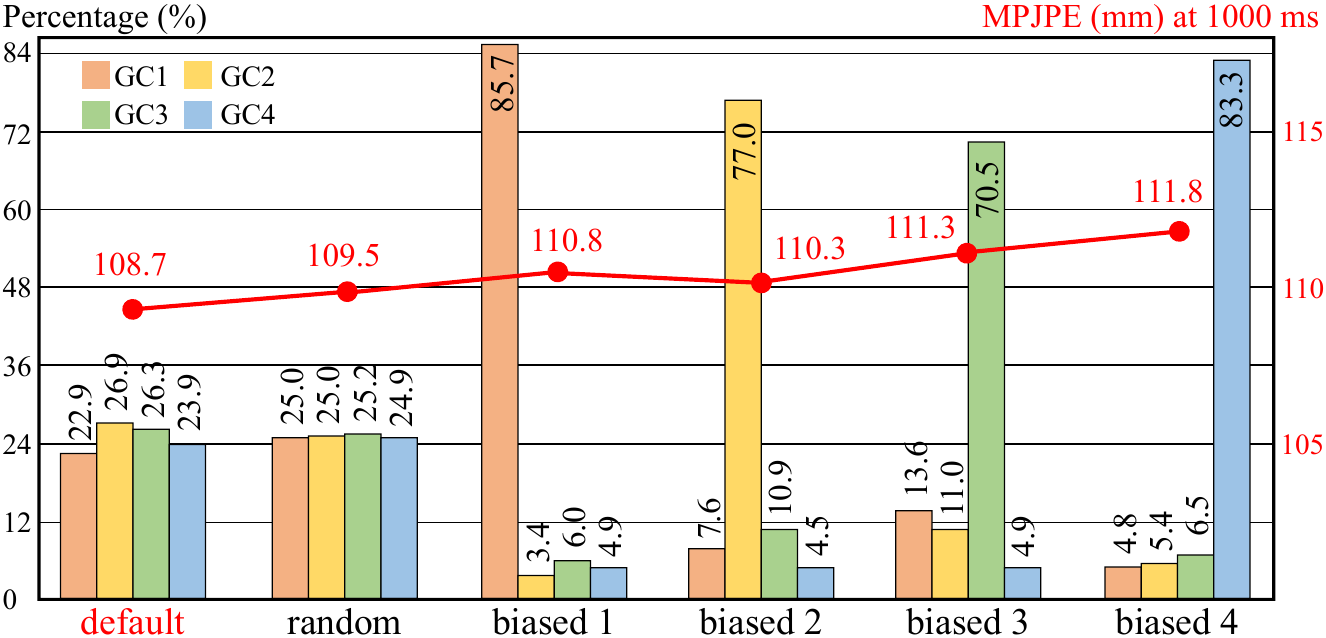}
    \caption{\small {Ablation on dynamic policy.}
    Six different policies (including unbiased, random, and biased towards GC 1--4) are compared, along with the corresponding MPJPEs at 1000ms (red line). The unbiased (default) policy achieves the best performance.
    }
    \label{fig: policy distribution}
    \end{figure}

We conduct ablation studies to verify two factors: 1) the necessity of a dynamic network over static ones; 2) the best architecture design within the dynamic network. Factor 1 is studied in Table \ref{ablation: dynamic}, and factor 2 in Figure \ref{fig: policy distribution}, Table \ref{ablation: graph convolution choices} \& \ref{ablation: block number}.

    \begin{figure}[t]
    \centering
    \includegraphics[width=\columnwidth]{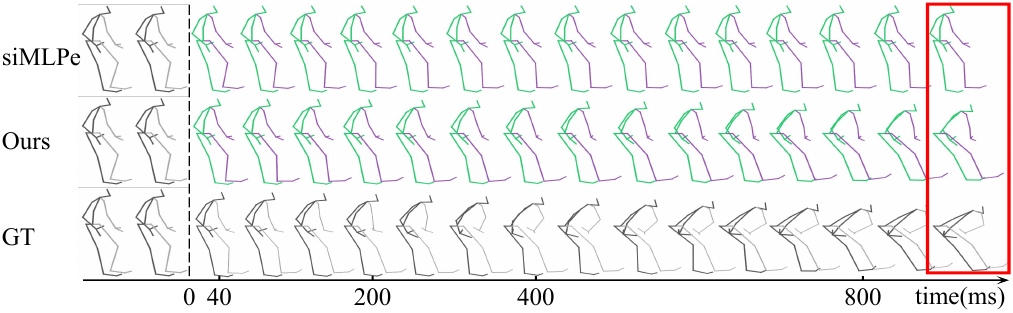} 
    \caption{\small \textbf{Qualitative results} of action ``sitting'' on H3.6M.
    }
    \label{viz}
    \end{figure}

\paragraph{\textbf{Necessity of Dynamic Network.}}
Table \ref{ablation: dynamic} verifies the advantages of our dynamic network compared against static networks.
Specifically, we design two types of static networks for ablation.
The first type, similar to existing GCNs, is built using one single type of graph convolution, which can be one of the special cases or the most general case of UniGC, i.e., 
$\textit{G}^\text{st},\textit{G}^\text{sc},\textit{G}^\text{tc},\textit{G}^\text{s},\textit{G}^\text{t},\textit{G}^\text{c},\text{or }\textit{G}$.
The second type adopts the original GCNext architecture, but instead of letting the selector select one at each layer, we simply use all graph convolutions by averaging or summing their outputs.
Table \ref{ablation: dynamic} shows that our GCNext (default) outperforms all static architectures, no matter how they are designed, verifying the necessity of a dynamic architecture for the task.

    \begin{table}[t]
    \centering
    \resizebox{\columnwidth}{!}{
    \begin{tabular}{l|lll|l|lll} \hline
    Arch. & 80 & 560 & 1000 & Arch.  & 80 & 560 & 1000 \\ \hline
    $\textit{G}^\text{st}$ & 9.8 & 75.8 & 109.9 & $\textit{G}^\text{s}$ & 10.1 & 76.5 & 110.7 \\
    $\textit{G}^\text{sc}$ & 9.6 & 76.0 & 109.9 & $\textit{G}^\text{t}$ & 10.1 & 77.0 & 111.0 \\
    $\textit{G}^\text{tc}$ & 11.0 & 80.9 & 113.6 & $\textit{G}^\text{c}$ & 10.2 & 77.8 & 111.6 \\
    $\textit{G}$ & 14.5 & 82.9 & 116.0           & all avg & 9.7 & 75.6 & 109.6 \\
    default & \textbf{9.3} & \textbf{74.7} & \textbf{108.7}                 & all sum & 10.1 & 77.1 & 110.9 \\ \hline
    \end{tabular}}
    \caption{\small \textbf{Ablation on dynamic architecture of GCNext}.
    The dynamic architecture of GCNext (default) is compared against different static variations.
    $\textit{G}^\text{st},\textit{G}^\text{sc},\textit{G}^\text{tc},\textit{G}^\text{s},\textit{G}^\text{t},\textit{G}^\text{c},\text{and }\textit{G}$ each represents a GCN built on a single type of graph convolution.
    ``all avg''/``all sum'' represents a GCN that uses multiple graph convolutions to average or sum their outputs, instead of selecting one.
    The results show that the dynamic architecture consistently achieves better performance than its static counterparts.
    }
    \label{ablation: dynamic}
    \end{table}

    \begin{table}[t]
    \centering
    \resizebox{\columnwidth}{!}{
    \begin{tabular}{m{0.4cm}<{\centering}m{0.4cm}<{\centering}m{0.4cm}<{\centering}m{0.4cm}<{\centering}m{0.4cm}<{\centering}m{0.4cm}<{\centering}|llll} \hline
    \multicolumn{6}{c|}{GC option} & \multirow{2}{*}{80} & \multirow{2}{*}{400} & \multirow{2}{*}{560} & \multirow{2}{*}{1000} \\
    $\textit{G}^\text{st}$ & $\textit{G}^\text{sc}$ & $\textit{G}^\text{tc}$ & $\textit{G}^\text{s}$ & $\textit{G}^\text{t}$ & $\textit{G}^\text{c}$ & & &   \\ \hline
    \checkmark & \checkmark & & & & & 9.8  & 57.1 & 75.2 &  109.5 \\
    \checkmark &  &  & \checkmark & & & 9.9  & 58.0 & 75.8 &  110.1  \\
    & \checkmark &  & \checkmark & & & 9.8  & 57.3 & 75.5 &  109.8 \\
    \checkmark & \checkmark &  &\checkmark & & & 9.8  & 57.2 & 75.3 &  109.7  \\
    
    \checkmark & & \checkmark & \checkmark & & \checkmark & 9.7  &  57.0  & 75.4 & 109.8  \\
    \checkmark & \checkmark & & \checkmark &  & \checkmark & 9.4  & 56.7 & 75.2 &  109.5  \\
    \checkmark & \checkmark & & \checkmark & \checkmark & & 9.5  & 56.9 & 75.2 &  109.1  \\
    \checkmark&\checkmark& & \checkmark &  &\checkmark & \textbf{9.3}& \textbf{56.4} & \textbf{74.7} & \textbf{108.7}  \\ \hline
    \end{tabular}}
    \caption{\small \textbf{Ablation on graph convolution options}.
    Each line represents a GCNext variant, whose option set is formed from all the graph convolutions marked with $\checkmark$.
    The best architecture is using $\{\textit{G}^\text{st},\textit{G}^\text{sc},\textit{G}^\text{s},\textit{G}^\text{c}  \}$ as the option set for the selector to select from.
    }
    \label{ablation: graph convolution choices}
    \end{table}

    \begin{table}[t]
    \centering
    \resizebox{\columnwidth}{!}{
    \begin{tabular}{l|llll|l|llll} \hline
    \# Layers   & 80 & 400 & 560 & 1000 & \# Layers & 80 & 400 & 560 & 1000\\ \hline
    12  &9.8  & 57.7  & 76.6 & 110.7    & 64          &  9.5   &  56.9   & 75.0  & 109.5  \\
    24  &9.6   & 57.2  & 75.3  & 109.8   & 80          & 9.5  &  57.9    & 75.3  & 109.6   \\
    48 & \textbf{9.3} & \textbf{56.4}  & \textbf{74.7} & \textbf{108.7}   &96   & 10.1 & 58.0 & 75.9 & 110.2 \\ \hline
    \end{tabular}}
    \caption{\small \textbf{Ablation on the number of layers}.}
    \label{ablation: block number}
    \end{table}

\paragraph{\textbf{Ablation on Graph Convolution Options.}}
At each layer, the selector is expected to select the best candidate graph convolution for each sample. While with more options comes higher representational capacity and more flexibility, training and optimization become more challenging. Therefore, we determine the optimal set of options through experiments. 
Specifically, we use different combinations of graph convolutions as different sets of options.
Table \ref{ablation: graph convolution choices} shows that the best architecture is providing the selector with 4 options: spatial-temporal ($\textit{G}^\text{st}$), spatial-channel ($\textit{G}^\text{sc}$), spatial ($\textit{G}^\text{s}$), and channel ($\textit{G}^\text{c}$) graph convolutions.

\paragraph{\textbf{Ablation on Number of Layers.}}
To determine the optimal model size, we study the influence of model size by decreasing or increasing the number of layers. Table \ref{ablation: block number} shows that GCNext with 48 layers yields the best performance.

\paragraph{\textbf{Ablation on Dynamic Policy.}}
We assess the impact of dynamic policy on GCNext in Fig. \ref{fig: policy distribution}.
Specifically, we compare 6 different policies, including unbiased (default), random, and biased towards different graph convolution types. GC 1--4 represent $\textit{G}^\text{st},\textit{G}^\text{sc},\textit{G}^\text{s},\text{ and }\textit{G}^\text{c}$ respectively, which are used in default GCNext architecture. The biased policy is obtained by piling more blocks of the desired type of graph convolutions on the original blocks such that the selector will make biased decision towards that type.
The red, yellow, green, and blue bars respectively represent the proportions of GC 1--4 being selected.
Fig. \ref{fig: policy distribution} shows that GCNext achieves the best performance with the unbiased policy.

\subsection{Visualization}
To put a finer point in the evaluation of our method, we present qualitative results, by visualizing the predicted motion sequence.
Fig. \ref{viz} provides a sample of action ``sitting'' on H3.6M. We compare the results of ours and of the most recent state-of-the-art, siMLPe~\cite{guo2023back}, along with the corresponding ground truth (GT). Our method generates more realistic and accurate body movements, as can be seen from the frame highlighted with red boxes.

\section{Conclusion}
This paper advances towards achieving unity of graph convolutions applied to 3D skeleton sequence data for human motion prediction. First, we propose Universal Graph Convolution (UniGC), which re-conceptualizing different graph convolutions, both existing and unexplored, as its special cases.
This innovation brings significant potential for discovering novel graph convolutions and evaluating diverse graph convolutions fairly. Moreover, we propose GCNext framework, which dynamically builds the optimal GCN architecture for each sample. GCNext offers multiple use cases, such as to be trained from scratch into a new GCN that better addresses the specific task, or to be used to refine a preexisting GCN.
We believe GCNext is a universal operator that can be applied to more than just human motion data. Exploiting its further application in more graph-structured spatial-temporal data types could be promising.

\section{Acknowledgments}
This work was supported by National Natural Science Foundation of China (No. 62203476), Natural Science Foundation of Shenzhen (No. JCYJ20230807120801002).

\bibliography{aaai24}

\end{document}